# Single Test Image-Based Automated Machine Learning System for Distinguishing between Trait and Diseased Blood Samples

Sahar A. Nasser[1], Debjani Paul[1], and Suyash P. Awate[2]

[1] Department of Bioscience and Bioengineering, Indian Institute of Technology Bombay (IIT-Bombay), India
[2] Department of Computer Science, Indian Institute of Technology Bombay (IIT-Bombay), India
* Senior Member, IEEE
** Fellow, IEEE

**Abstract—** We introduce a machine learning-based method for fully automated diagnosis of sickle cell disease of poor-quality unstained images of a mobile microscope. Our method is capable of distinguishing between diseased, trait (carrier), and normal samples unlike the previous methods that are limited to distinguishing the normal from the abnormal samples only. The novelty of this method comes from distinguishing the trait and the diseased samples from challenging images that have been captured directly in the field. The proposed approach contains two parts, the segmentation part followed by the classification part. We use a random forest algorithm to segment such challenging images acquitted through a mobile phone-based microscope. Then, we train two classifiers based on a random forest (RF) and a support vector machine (SVM) for classification. The results show superior performances of both of the classifiers not only for images which have been captured in the lab, but also for the ones which have been acquired in the field itself.

**Index Terms—** Sickle Cell Disease, Microscope, Image, Low contrast, Machine Learning, Segmentation, Random Forest, Support Vector Machines, Shape Descriptors, Roundness, Form Factor, Solidity.

## I. INTRODUCTION

Sickle cell disease is a monogenetic disorder transferred from the parents to the off-springs [1, 2]. This disease is caused by a mutation occurs in the β-globin gene which leads to the formation of polymeric chains inside the red blood cell (RBC) [18]. Polymerization deforms the shape of the RBC and as a result it losses its membrane flexibility. The sickle cells stick on the inner walls of the vessels and then block them [1, 3]. If both of the parents are carriers of the disease, some of their children may get sickle cell disease. According to 2011 population census, the population of India is 10.45 crore of these 1crore are carriers of the sickle cell disease and 14 lakhs have the disease [16]. One of the major reasons for the prevailment of this disease in the developing countries is that couples before marriage do not have access to tests and hence do not know their status, due to lack of equipment and trained personnel that are necessary for the test. Hence, in order to alleviate this major health crisis, there is a huge need for a cheap and a portable device which can be used directly in the field by a local health care worker. As an accurate prediction can not only prevent the birth of new diseased people, it also saves the lives of many carriers who are not aware of their situations, as the carrier may face the same symptoms of the diseased in the extreme conditions which may lead to death. To meet this need our team has developed a mobile phone-based microscope to image the whole blood samples and study their behaviors in a microfluidic channel. While the device is cheap and affordable, the demand for an expert microscopist to analyze the images and make decisions about them still exist. To tackle this complication, we developed a machine learning-based program that is able to process the images and give a decision about the sample with testing accuracies reached 100% for both the SVM classifier and the RF one.

Various researches have been done to classify images of a variety of cells into normal and abnormal. Some of these investigations are based on conventional image processing techniques such as morphological operations which require high-quality images with high contrast between the cells and the background, and necessitate user interaction as well [4]. Also, some researchers have built machine learning-based algorithms to classify high-quality microscopic images of a diversity of cells. For instance, Velchamy et al used the morphological operations to detect the RBCs which have sufficient contrast from the background [5]. Twenty-seven geometrical and statistical features were used as an input of a neural network which classified the samples into normal and abnormal only, with classification efficiency 80% and 66.6% respectively. Other machine learning based algorithms like the RF classifier combined with K-mean clustering have been proposed by Essa and his colleagues to segment the human osteosarcoma phase contrast images with precision 92.96% and recall 96.69% [6]. Many other groups exploited the fact that the detected cells have nuclei which make the task to segment such cells much easier [19].

To our knowledge, till now none of the previous researches has differentiated the trait red blood samples from the diseased ones. In this paper, we presented a robust algorithm which deals with low-quality mobile phone images and gives a prediction whether a sample is normal, trait, or diseased. This work is divided into two parts. The first part focuses on developing a segmentation algorithm to extract the features of RBCs from low quality mobile phone images. While the second part of the paper discusses the classification algorithm, selection of the best geometrical features to use, and optimization of the hyperparameters of the implemented classifiers which are the RF classifier and the SVMs one to get the optimum performances, and eventually comparison of the executions. Finally, we used our method to analyze images acquired in the field, which are more challenging than the ones which captured in the lab, and we got satisfied results. We believe that our method is superior to those described previously for applications in poor rural settings where high-quality microscopes might not be available.

## II. PROPOSED METHOD

### A. EXPERIMENT

2ml of venous blood samples was collected from participants as a part of the sickle cell screening program by Dayanand Hospital, Talasari. 0.5ml of the collected blood was used for the sickle cell assay. Experiments were performed with two concentrations (0.1% and 0.3%) of oxygen scavenger (Sodium metabisulphite). 5μl of oxygen scavenger was mixed with 95 μl of blood (20X dilution) and 10μl of this was loaded onto the microfluidic device for imaging. Image was captured for 0min and 30min. Important factors like temperature and ambient lighting varied in the field as compared to in the laboratory. Variation in temperature was recorded throughout all the experiments (23-32℃).

### B. SEGMENTATION

In our proposed method we use shape descriptors such as roundness and form factor to distinguishing the normal, trait, and diseased samples. To extract the above-mentioned descriptors, we need to segment the cells from the background as a preceding step. Many researchers who tackled similar problems such as [7, 8, 9] used morphological processes namely thresholding, opening, closing, and so on. Others like in [10] balanced the average values of the red , green, and blue channels under dissimilar illuminations followed by geometrical filtering. While Veluchamy et al used edge detection-based morphology tools to segment the images [5]. Dewan et al [19], presented an algorithm to detect the biological cells by applying top-hat filter followed by Gaussian filtering and h-maxima transformation to segment the nuclei of the cells which had good contrast from the background. However, all the previously indicated methods required good quality of images with high contrast and minimum or no illumination changes, and some of them exploited the fact that the nucleus of the cell has good contrast with the background. In contrast, our *mobile microscopic images* suffer from low quality, as the mobile phone saves the images in jpeg format , the low contrast between the cells and the background, and the radical changes in the illumination due to many factors such as manufacturing problems of the hardware, using different mobile phones, different circumstances (surrounding light inside the lab vs. in the field) and skill-based mistakes of the staff, see Figure 2 (a, d, g, j, m, p). Not only that, the absence of nuclei of the RBCs makes these images much more challenging to be segmented. And the crenation of some samples after inducing the sickling increases the difficulty of this task.

In our proposed method we use one of the machine learning algorithms, the random forest, to achieve the segmentation task. We start with describing of the suggested segmentation algorithm followed by the results.

**Random Forest Algorithm :**

Every random forest consists of many decision trees. For each tree, during inference, the data flow from the root through many nodes till the leaves where the final decision is made. Splitting the data distribution at each node follows the entropy reduction principale, which aims to the increase information gain[11, 12]. In our case we have three classes where every pixel in the image is either inside the cell, or on the boundary, or belongs to the background. Let us consider a forest with T trees. Then, splitting the data distribution at a node j of a tree t∈{1, 2, ..., T}, requires choosing a threshold $\theta_j^*$ which gives us the highest information gain

$$\theta_j^* = argmax_{\theta_j \in \tau_j} I_j \qquad (1)$$

where τ is the threshold parameter for splitting. The prediction of a tree t is the average of its leaves' predictions, and the final prediction of the forest is the average of its trees' predictions. The segmentation method contained three phases: the training phase, the validation phase for optimizing the free parameters of the forest by implementing cross validation, and the testing phase for testing the performance of the classifier. We used 70% of the data for training and 30% used for validation. The segmentation results were imperfect before tuning the free parameters of the classifier. However, we experimentally proved that many parameters of the RFC had no significant effect on its performance. As [11, 13] referred to, there are only key two free parameters of the RF to tune, which are the size of the forest and the depth of its trees. We found that the more we increase the size of the forest, the more we improve in the quantitated accuracy and in the visual appearance of the segmented image. But, we have to take into account the size of the training set and the increment of the computational cost as well. The second parameter is the depth of the trees, in other words, the level upto which the trees are allowed to expand. We noticed that for shallow trees some artifacts appeared in the segmented image. Hence, increasing the depth to a certain level improved both of the accuracies and the visual appearance as well. However, increasing the depth above the optimum value produced some artifacts in the resulting segmentation, see. Furthermore, there is one more parameter which has a major influence on the accuracies and the quality of the segmentation, which is the length of the input features vector. Larger patch sizes can increase information content in the feature, but can also make the learning more difficult in a higher dimensional space that would entail larger training sets. Due to memory limitations, we cannot train the classifier on a huge dataset. It experimentally emerged that a patch of size (21×21) is the best choice for our images as bigger patches caused some artifacts in the images due to the overfitting. The final results reached 0.965 for training accuracy and 0.91 for validation one.

### C. CLASSIFICATION

As we have mentioned before, the input of the classifiers are either the distribution of the roundness or the distribution of the form factor of the input image. To get the distributions of the cell features from the segmented image, we applied the connect component analysis on the segmented image followed by some morphological operations to fill the holes and subtract the background. Then we computed the solidity of every cell of the input image, and we exclude the ones, which have a solidity less than 0.8, from contributing in the distributions of the roundness and form factor. We explained the random forest algorithm in-depth in the segmentation section; the only difference while applying the RF for classification are the input features to the forest, and the classes of the output which are for this part normal, trait, and diseased. Hence, in this section, we focus on describing the SVMs based algorithm for classification.

**Support Vector Machines:**

Support vector machines (SVMs) is one of the most powerful supervising learning methods. It has been used not only for the linear classification but as well for nonlinear classification via kernelized support vector machines [14]. Our problem is a non-linear classification problem and we use kernelized SVMs which take the original input data space and transform it into higher dimensional feature space, where it becomes easier to classify the transformed data using a linear classifier. There are various kernels such as the radial basis function kernel (RBF), the polynomial kernel, the sigmoid kernel and so on. We found that the best kernel to tackle our classification problem is the RBF [15].

$$K(x, x_i) = exp[-\gamma \cdot \|x - x_i\|^2]$$

The kernel measures the similarity between the input data. The free parameter for of this kernel is gamma. Another free parameter of the SVM is the regularization parameter C [15]. Where small γ means a larger similarity radius which indicates that the points that are far apart from each other are considered to be similar, that means more points group together, which produces smoother decision boundaries. On the other hand, a large gamma causes smaller similarity radius which leads to a rapid decay of the decision boundary, Hence the points need to be close to each other to consider similar. While the regularization parameter C controls the tradeoff between satisfying maximum margin criterion, to find a simple decision boundary, and avoiding the misclassification errors on the training set.

## III. RESULTS AND DISCUSSION

The whole dataset contained 28 diseased, 91 normal and 37 trait samples. 5 diseased, 7 trait and 15 normal samples were kept aside for testing the performances of the classifiers on unseen data, from the remaining 129 images 70% were used for training and the rest 30% for validation. We obtained the tags of our images from high performance liquid chromatography which we got from Valsad Raktdan Kendar Hospital. First of all, we compared the performances of the classifiers for two different input features the roundness, and the form factor. As [8] stated the form factor was the best feature to classify red blood samples into normal and abnormal. However, we obtained the best performances of the classifiers for the roundness feature as an input, which was the most effective to distinguish between the three classes. Next step was tuning the free parameters of the classifiers. The optimum parameters of the classifiers were for RF 100 trees as the size of the forest, depth equals three and maximum number of features at each node equals $\sqrt{N}$. For the SVM the optimum parameters were RBF kernel with C=250 and gamma= 1. The final results of both classifiers were 100% for the sensitivity, the specificity, and the accuracy for the testing samples 27 samples which have been captured in the lab. The previous results proved that there are significant differences in the shapes of the cells belong to the diseased samples at concentration 0.1 of the oxygen scavenger after 30 minutes of inducing the sickling and the trait ones at concentration 0.3 of it after inducing the sickling. We went further and we tested our method to diagnose samples captured in the field itself. In the field, as we do not have previous information whether the sample belongs to a diseased person or a trait one, the protocol which we follow in such cases is to take two samples from each subject and treat one of them with 0.1 concentration of the oxygen scavenger and the other with 0.3. The classifier will take the decision according to the following discussion:

Assume $p_1$ is the prediction of the classifier about the sample at concentration 0.1, and $p_2$ is its prediction about the sample at concentration 0.3. And 0, 1, and 2 are the labels of sickled, trait, and normal classes respectively. Then the final prediction of the classifier about the subject will be one of the following situations:

If $(p_1 = 0$ and $p_2 = 0)$ or $(p_1 = 0$ and $p_2 = 1)$ or ($p_1 = 1$ and $p_2 = 1)$ then the final decision is a diseased subject.

If $(p_1 = 2$ and $p_2 = 0)$ or $(p_1 = 2$ and $p_2 = 1)$ or $(p_1 = 1$ and $p_2 = 0)$ then the final decision is a trait subject.

If $(p_1 = 2$ and $p_2 = 2)$ then the prediction is a normal subject.

We tested our algorithm on 29 samples three of them were diseased and the rest were trait. The results were as shown in Table(1). However, the dataset size was very small and, while the current results are very promising, further evaluation is needed on larger datasets to test the efficacy of the method.

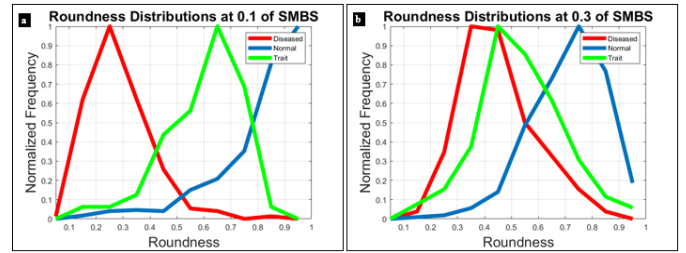

Fig. 1. The distribution of the roundness after 30 minutes of inducing the sickling for three samples (trait, diseased, and normal) at 0.1 concentration of oxygen scavenger as in (a) and three samples belong to the same subjects at a concentration 0.3 of oxygen scavenger as in (b).

Table 1. The effects of Form Factor and the Roundness on the performances of the classifiers.

| Feature | Form Factor | | Roundness | |
| --- | --- | --- | --- | --- |
| *Classifier* | RF | SVM | RF | SVM |
| *Accuracy* | 52.6% | 72% | 96.5% | 93% |
| *Sensitivity* | 100% | 67% | 100% | 67% |
| *Specificity* | 43.7% | 72.7% | 96% | 96% |

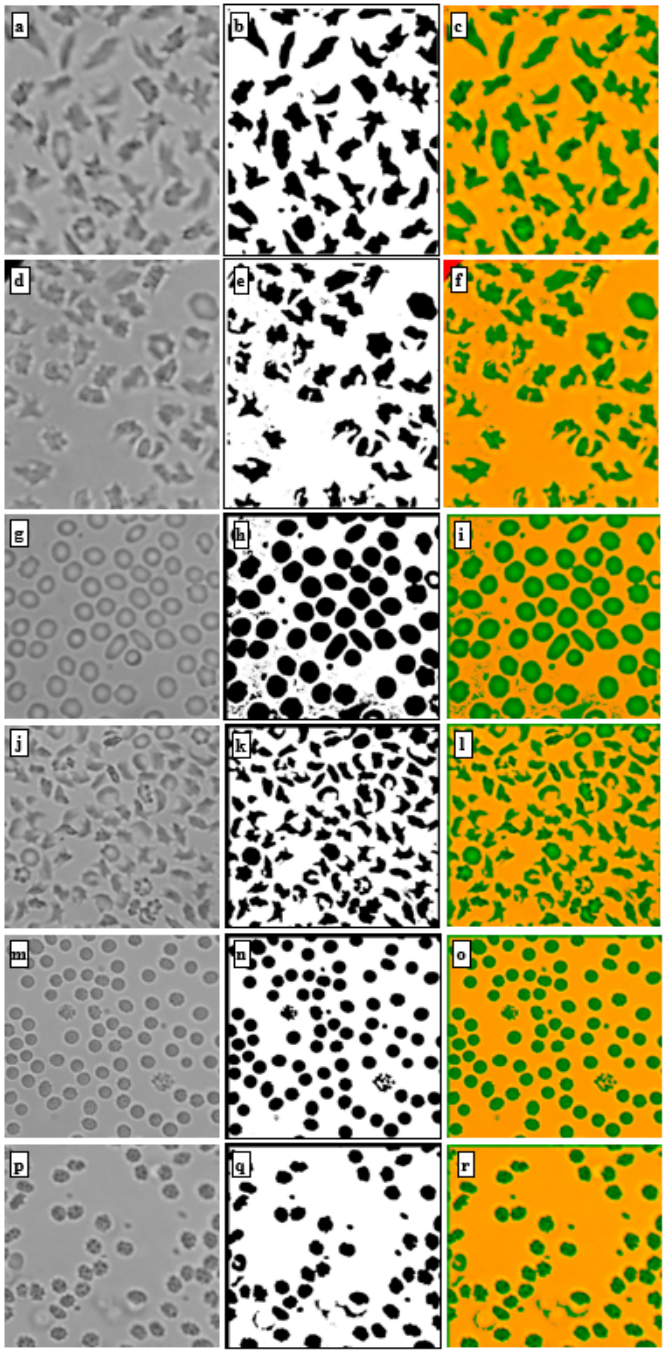

Fig. 2. The results of the segmentation algorithm. The images in this figure are patches belong to the samples reported in Figure.1. Images (a, d, g, j, m, p) are patches belong to the samples diseased at concentration 0.1, diseased at concentration 0.3, trait at concentration 0.1, trait at concentration 0.3, normal at concentration 0.1, normal at concentration 0.3 respectively. Images (b, e, h, k, n, q) are the segmented images belong to the patches (a, d, g, j, m, p) respectively. Images (c, f, i, l, o, r) are the results of overlaying every segmented image (b, e, h, k, n, q) on top of its original patch (a, d, g, j, m, p). The experiments have been done for optimum values of the free parameters (forest size= 50 trees, depth= 5, maximum number of features at each node= $\sqrt{N}$ where N is the total number of features, patch size 21×21).